\documentclass[10pt,journal,compsoc]{IEEEtran}
\usepackage[utf8]{inputenc}
\usepackage[T1]{fontenc}
\usepackage{hyperref}
\usepackage{url}
\usepackage{booktabs}
\usepackage{amsfonts}
\usepackage{nicefrac} 
\usepackage{microtype} 
\usepackage{graphicx}
\usepackage{comment}
\usepackage{amsmath,amssymb}
\usepackage{color}
\usepackage{multirow}
\usepackage{cases}
\usepackage{wrapfig}
\usepackage{lipsum}
\usepackage{subfig, lipsum}
\usepackage{array}
\usepackage[numbers,sort&compress]{natbib}
\usepackage[usestackEOL]{stackengine}
\strutlongstacks{T}

\usepackage{floatrow}
\newfloatcommand{capbtabbox}{table}[][\FBwidth]

\def\etal{\emph{et al.~}}
\def\ie{\emph{i.e.}}
\def\eg{\emph{e.g.}}

\begin{document}

\title{Towards Visually Explaining Similarity Models}

\author{Meng Zheng, \IEEEmembership{Member, IEEE,}
	Srikrishna Karanam, \IEEEmembership{Member, IEEE,}
	Terrence Chen, \IEEEmembership{Senior Member, IEEE,}
	Richard J.~Radke, \IEEEmembership{Senior Member, IEEE,} and
	Ziyan Wu, \IEEEmembership{Member, IEEE}
	\IEEEcompsocitemizethanks{\IEEEcompsocthanksitem M.~Zheng, S.~Karanam, T.~Chen, and Z.~Wu are with United Imaging Intelligence, Cambridge, MA 02140 USA (e-mail: \{first.last\}@united-imaging.com). Corresponding author: S. Karanam.}% <-this % stops a space
	\IEEEcompsocitemizethanks{\IEEEcompsocthanksitem  R.J.~Radke is with the Department of Electrical, Computer, and Systems Engineering, Rensselaer Polytechnic Institute, Troy, NY 12180 USA (rjradke@ecse.rpi.edu).}% <-this % stops a space
	}
	
\IEEEtitleabstractindextext{%
\begin{abstract}
We consider the problem of visually explaining similarity models, \ie, explaining why a model predicts two images to be similar in addition to producing a scalar score. While much recent work in visual model interpretability has focused on gradient-based attention, these methods rely on a classification module to generate visual explanations. Consequently, they cannot readily explain other kinds of models that do not use or need classification-like loss functions (\eg, similarity models trained with a metric learning loss). In this work, we bridge this crucial gap, presenting a method to generate gradient-based visual attention for image similarity predictors. By relying solely on the learned feature embedding, we show that our approach can be applied to any kind of CNN-based similarity architecture, an important step towards generic visual explainability. We show that our resulting attention maps serve more than just interpretability; they can be infused into the model learning process itself with new trainable constraints. We show that the resulting similarity models perform, and can be visually explained, better than the corresponding baseline models trained without these constraints. We demonstrate our approach using extensive experiments on three different kinds of tasks: generic image retrieval, person re-identification, and low-shot semantic segmentation.
\end{abstract}
}

\maketitle
\IEEEdisplaynontitleabstractindextext

\IEEEraisesectionheading{
\section{Introduction}
\label{sec:intro}}
\begin{figure*}[!h]
	\centering
	\includegraphics[draft=false,width=1\linewidth]{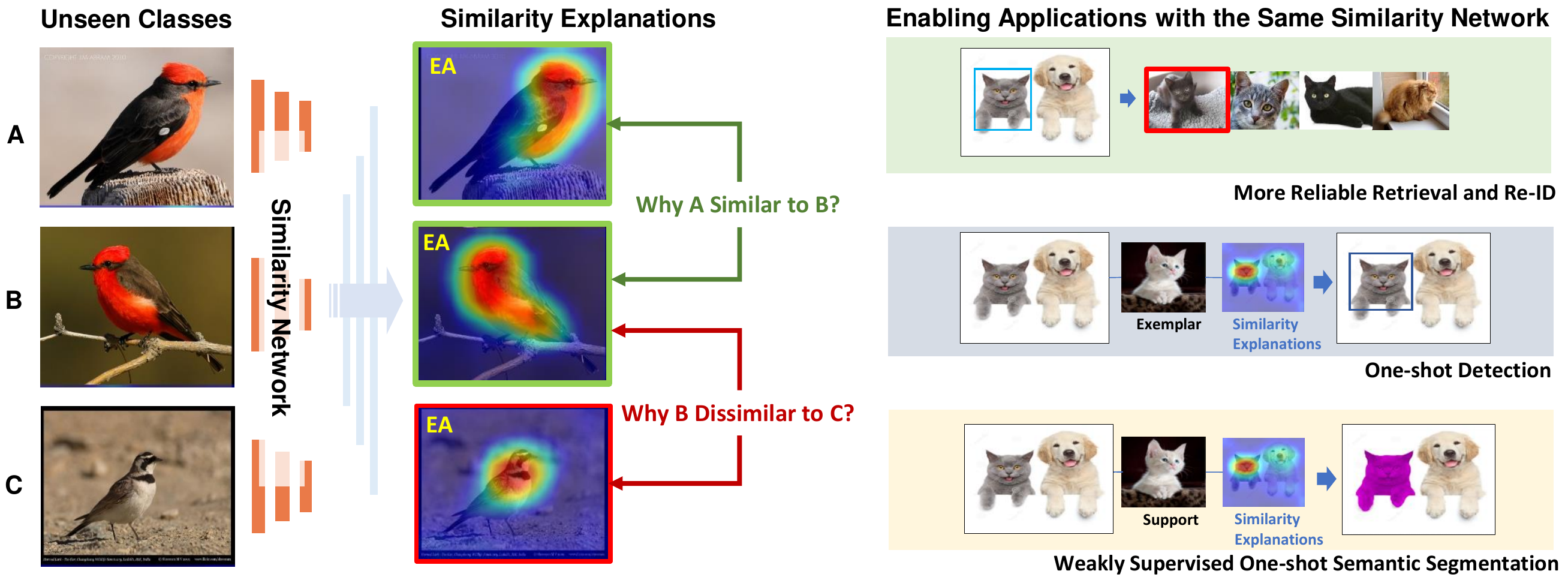}%
	\caption{We propose to visually explain similarity models, showing our method's versatility in diverse applications such as metric learning and semantic segmentation.} 
	\label{fig:teaser}
\end{figure*}
As we work towards transitioning algorithmic deep learning research to real-world applications, we must consider the important question of model explainability and interpretability. This is especially critical for applications such as healthcare where medical professionals may not readily trust the decisions of a black-box artificial intelligence system for diagnostic purposes. 

Following the visualization work of Zeiler and Fergus \cite{zeiler2014visualizing} and Mahendran and Vedaldi \cite{mahendran2015understanding}, much recent progress in visually explaining convolutional neural networks (CNNs) has been led by attention-based techniques \cite{CAM_CVPR16,GradCAM_ICCV17} that produce attention maps highlighting regions in input images that are considered (by the model) to be important for the final prediction. GradCAM \cite{GradCAM_ICCV17} has been particularly impactful with its simple and intuitive attention generation mechanism as well as extensibility in using the resulting attention to enforce trainable constraints \cite{GAIN_CVPR18,wang2019sharpen}. While ``attention'' may have different connotations \cite{bahdanau2015neural,andreas2016neural,residualAtt_CVPR17,AttentionIA_NIPS17}, in our work, we refer to attention computed by means of the gradient of a differentiable activation in the spirit of GradCAM. 

A key limitation of GradCAM, and other related methods, is that they rely on activations from a classification module (\eg, a fully-connected unit followed by softmax) to generate the attention map. While there are many models in computer vision that rely on a classification loss function during training, there are even more that do not need one (\eg, generative models, autoencoders, similarity models). While it is desirable to explain a wide variety of models (not just classification ones), extending  gradient-based explanation methods to these models is not trivial. Furthermore, adding a classification loss function just for the sake of computing attention maps to (for example) a similarity model that is typically trained with a metric learning loss seems suboptimal. Consequently, the question we ask is: \textsl{can we generate visual attention maps for models without needing a classification module for computing visual attention?} Here, we address this question in the context of similarity models trained on image data.

\begin{figure}[!h]
	\centering
\includegraphics[draft=false,width=\linewidth]{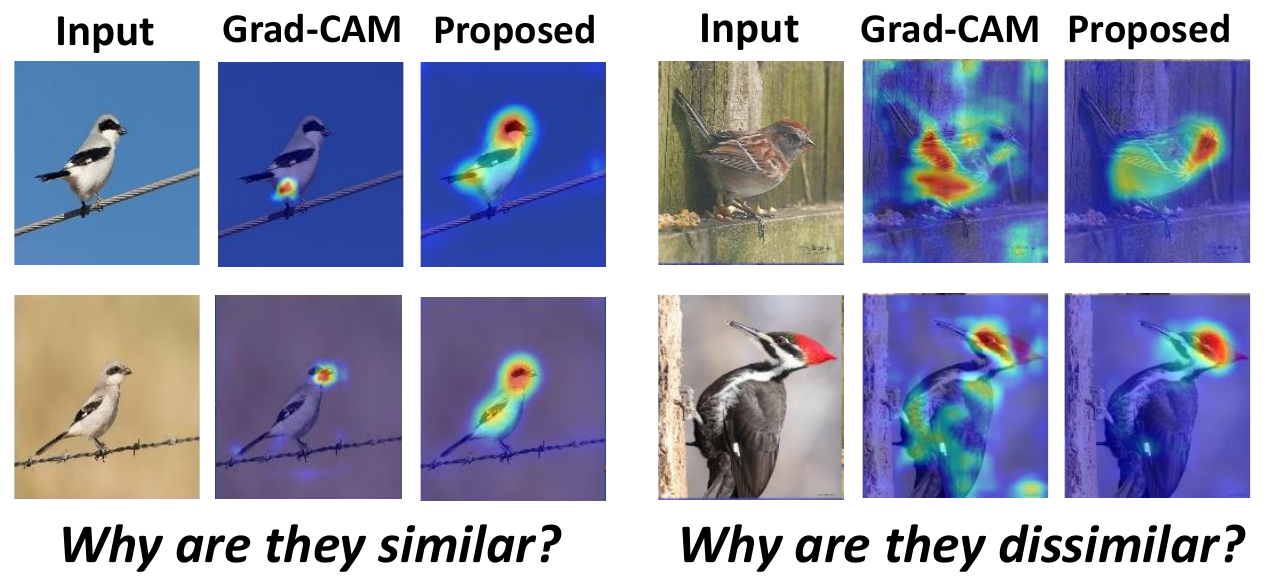}%
\caption{\label{fig:gcam} GradCAM vs. proposed method. One can note our method is able to highlight corresponding regions more clearly when compared to GradCAM.}
\end{figure}

A similarity model is trained to embed same-category images close together in the learned feature space while those of different categories farther apart. This is typically achieved using a metric learning loss. Our definition of a visual explanation for such a model is a set of attention maps that highlights regions in the input images that represent the model's evidence that the images are similar, \ie, close in the learned feature space. For instance, in the case of a Siamese similarity model (trained with pairs of data), we will generate two attention maps highlighting regions that the model reasons are most important for a particular pair to be similar. Similarly, for a triplet similarity model (trained with triplets of data), we will generate three attention maps that highlight regions that the model reasons are most important for this particular input to satisfy the triplet condition (i.e., the anchor image is close to the positive image but far from the negative image). We refer to such sets of attention maps as \textsl{\textbf{similarity attention}} (see Fig.~\ref{fig:teaser}). 

As noted above, one can certainly add a classification module to a similarity model so that the resulting classification activations can be used to generate attention maps with GradCAM. In Fig.~\ref{fig:gcam}, we show the result of this operation (this Siamese model was trained with the contrastive and binary cross entropy loss), where we see the GradCAM attention maps highlighting non-corresponding (red) regions for the similar pair whereas our proposed similarity attention more clearly highlights corresponding regions. Comparing these two figures, it is clear that simply adding a classification term so one can use GradCAM to generate attention does not result in the kind of intuitively satisfying attention maps one would expect; our proposed method explicitly bridges this gap.

Our method does not need a classification module to generate attention maps, thereby addressing a key limitation of GradCAM. Furthermore, as shown in Fig~\ref{fig:gcam}, our method produces more intuitive attention maps for similarity models, thereby removing the need for training the similarity model with a classification objective just to compute attention maps. Instead, our approach is based on identifying feature dimensions that are important for two similar images to be embedded close in the feature space. Starting from such analysis, we propose to generate a differentiable activation that can be used to compute gradients with respect to convolutional feature maps and hence attention maps, or similarity attention as noted above. A useful by-product of this approach, leading to our next contribution, is that these attention maps result in explicit constraints, called \textsl{\textbf{similarity mining}}, for further bootstrapping the learning of the similarity model, which we demonstrate results in improved downstream performance (\eg, better rank performance) as well as attention maps.  

A few recent examples of attempts to visually explain similarity models include Plummer \etal \cite{whymatch_2019} and Chen \etal \cite{Chen_WACV20}. However, as detailed in Section~\ref{sec:relatedWork}, our method is more flexible and generic in several aspects. Briefly, while Plummer \etal \cite{whymatch_2019} needs attribute labels coupled with an attribute classification module and a saliency generator to generate explanations, our method removes this need for extra labels (i.e., we only need yes/no pairwise labels) or additional learning modules (i.e., we do not need attribute classifiers). While Chen \etal \cite{Chen_WACV20} addresses the need for attribute information by relying on the learned feature space, instead of explaining why two testing images are similar, it ends up explaining why the two training images closest to the test images are similar by means of a training nearest neighbor search database. 

In order to demonstrate technical generality, we show how our method can be used to generate attention maps for a variety of similarity architectures (\eg, Siamese, triplet, quadruplet). We do this both theoretically, deriving raw attention matrices conditioned on the learned feature space, and empirically. In order to demonstrate wide applicability, we conduct experiments on three different tasks: generic image retrieval, person re-identification (re-id), and low-shot semantic segmentation. While retrieval and re-id are standard metric learning applications, the segmentation application shows how our proposed method can do much more than just help explain why two images are similar. Specifically, we demonstrate how these cues can help to discover corresponding regions of interest (in this case between a query and a support image) and perform semantic segmentation, as overviewed in Fig.~\ref{fig:teaser}.

\section{Related Work}
\label{sec:relatedWork}
Our work is related to both visual explainability and similarity/metric learning, and we briefly review closely-related methods along these directions, helping differentiate our work and put it in proper context.

\subsection{Learning Visual Explanations}
Dramatic performance improvements of vision algorithms driven by black-box CNNs have led to a recent surge in attempts to explain and interpret model decisions \cite{zeiler2014visualizing,mahendran2015understanding,CAM_CVPR16,residualAtt_CVPR17,GradCAM_ICCV17,AttentionIA_NIPS17,gradCAMpp_WACV18,ABN_CVPR19}. To date, most CNN visual explanation techniques fall into either response-based or gradient-based categories. Class Activation Map (CAM) \cite{CAM_CVPR16} used an additional fully-connected unit on top of the original deep model to generate attention maps, thereby requiring architectural modification during inference and limiting its utility. Grad-CAM \cite{GradCAM_ICCV17}, a gradient-based approach, solved this problem by generating attention maps using class-specific gradients of predictions with respect to convolutional layers. Some follow-up work \cite{GAIN_CVPR18,wang2019sharpen,zheng2019re} then took a step forward, using the attention maps to enforce trainable attention constraints, demonstrating improved model performance. These methods, however, were specific in their focus. While Li \etal \cite{GAIN_CVPR18} and Wang \etal \cite{wang2019sharpen} focused on categorization, Zheng \etal \cite{zheng2019re} focused on re-id, leading to application-specific assumptions (e.g., upright standing persons in re-id) and pipeline design (e.g., needing a classification module to compute attention). In our work, we propose a \textsl{generic} method that can generate similarity attention from, in principle, any similarity measure that relies on computing distances between two or more feature vectors. Additionally, by design, our technique can facilitate the generation and enforcement of trainable constraints with the resulting similarity attention maps, leading to improved model performance. 

A few recent methods \cite{Chen_WACV20, whymatch_2019} presented techniques for explaining deep similarity neural networks. Given a pre-trained image similarity embedding model that is to be (visually) explained, Plummer \etal \cite{whymatch_2019} proposed to learn an attribute predictor and an associated saliency generator. The attribute predictor, implemented using a multi-class classifier with ground-truth attributes, predicted the attribute of interest which was then explained using its associated saliency map. While this method relies on classification predictions, in addition to training more models (attribute predictor and saliency generator along with original similarity model), our method is more flexible in that it does not need any extra training or attribute labels, while also removing the need for a classification module. 

Chen \etal \cite{Chen_WACV20} took a step in the direction of reduced dependence on classification by proposing a two-step technique to explain a similarity embedding network. During training, a database of GradCAM-like gradients (grad-weights) was created with training data, which was then queried during testing to determine the grad-weight, and hence the attention map, for each test image. While this may seem reasonable given the lack of classification logits in a similarity model, what this approach is doing in essence is explaining why the two closest training images (from the database) are similar, but not the two testing images. Other potential issues include the dependence on accurate nearest neighbor search, which may be challenging in cases of a significant domain shift, thus limiting its extensibility to other applications. Our proposed method addresses some of these issues by design: since it does not have any nearest neighbor look-up, it in principle explains why two images are similar. Furthermore, due to its dependence on only the learned embedding, a model trained with our method has the potential to generalize and be used (even without retraining if needed, as we show later on) in applications that are quite different from metric learning/retrieval.

\subsection{Learning Distance Metrics}
Metric learning approaches attempt to learn a discriminative feature space to minimize intra-class variations, while also maximizing the inter-class variance. Traditionally, this translated to optimizing learning objectives based on the Mahalanobis distance function or its variants \cite{LMNN_NIPS09,KISSME_CVPR12,LFDA_CVPR13,LOMO_XQDA_CVPR15}. Much recent progress with CNNs has focused on developing novel objective functions or data sampling strategies. Wu \etal \cite{wu2017sampling} demonstrated the importance of careful data sampling, developing a weighted data sampling technique that resulted in reduced bias, more stable training, and improved model performance. On the other hand, Harwood \etal \cite{SmartMF_ICCV17} showed that a smart data sampling procedure that progressively adjusts the selection boundary in constructing more informative training triplets can improve the discriminability of the learned embedding. 

Substantial effort has also been expended in proposing new objective functions for learning the distance metric. Some recent examples include the multi-class N-pair \cite{Npairmc_NIPS16}, lifted structured embedding \cite{LiftStruct_CVPR15}, proxy-NCA \cite{ProxyNCA_ICCV17} and Fast-AP \cite{FastAP_CVPR19} losses. The goal of these and related objective functions is essentially to explore ways to penalize training data samples (pairs, triplets, quadtruplets, or even distributions \cite{rippel2015metric}) so as to learn a discriminative embedding. In this work, we take a different approach. Instead of just optimizing a distance objective (\eg, triplet), we also explicitly consider and model network attention during training. This leads to two key innovations over existing work. First, we equip our trained model with decision reasoning functionality. Second, by means of trainable attention, we guide the network to discover local regions in images that contribute the most to the final decision, thereby improving model performance which we demonstrate empirically. 

\begin{figure*}[!h]
	\centering
	\includegraphics[draft=false,width=1\linewidth]{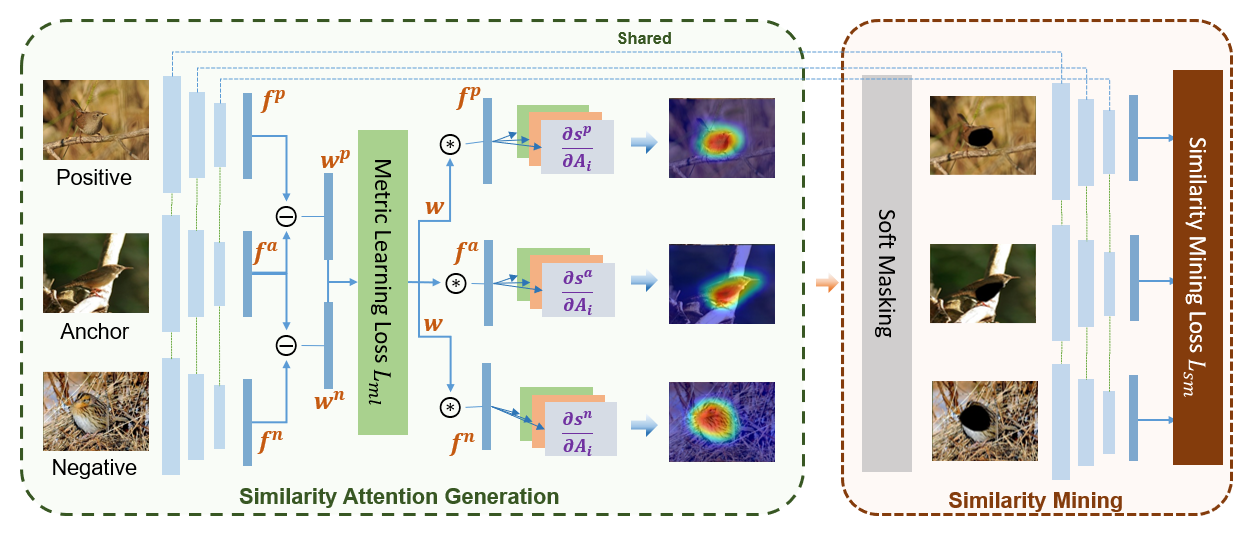}%
	\caption{Overall pipeline of our similarity attention generation and mining technique.} 
	\label{fig:pipeline}
\end{figure*}

\section{Method}

Given $N$ labeled images $\{(\mathbf{x}_{i},y_{i})\}$ each belonging to one of $k$ categories, where $i=1,\ldots,N$, $\mathbf{x} \in \mathbb{R}^{H\times W\times c}$, and $y \in \{1,\ldots,k\}$, we propose a method to visually explain why a similarity model predicts two images $\mathbf{x}_{1}$ and $\mathbf{x}_{2}$ to be similar. In Section \ref{sec:SA}, we discuss our proposed technique to generate these visual similarity explanations and show how it can be easily integrated with existing similarity models. In Section~\ref{sec:SA_mining}, we discuss how our proposed attention generation mechanism facilitates principled training of similarity models with our new similarity mining learning objective.

\subsection{Generating visual similarity explanations}
\label{sec:SA}
Traditional similarity predictors such as Siamese or triplet models are trained to respect the relative ordinality of distances between data points. For instance, given a training set of triplets $\{(\mathbf{x}^{a}_{i},\mathbf{x}^{p}_{i},\mathbf{x}^{n}_{i})\}$, where $(\mathbf{x}^{a}_{i},\mathbf{x}^{p}_{i})$ have the same categorical label while $(\mathbf{x}^{a}_{i},\mathbf{x}^{n}_{i})$ belong to different classes, a triplet similarity predictor learns a $d-$dimensional feature embedding of the input $\mathbf{x}$, $f(\mathbf{x}) \in \mathbb{R}^{d}$, such that the distance between $f(\mathbf{x}^{a}_{i})$ and $f(\mathbf{x}^{n}_{i})$ is larger than that between $f(\mathbf{x}^{a}_{i})$ and $f(\mathbf{x}^{p}_{i})$ (within a predefined margin $\alpha$). This is typically achieved by minimizing the triplet loss (with pre-defined margin $m$) over the set of K triplets $\{(\mathbf{x}_{i}^{a},\mathbf{x}_{i}^{p},\mathbf{x}_{i}^{n})\}, i\in \{1, 2, \ldots, K\}$:

\begin{equation}
    L_{ml}=\frac{1}{K}\sum_{i=1}^{K}\text{max}(\|f(\mathbf{x}^{a}_{i})-f(\mathbf{x}^{p}_{i})\|-\|f(\mathbf{x}^{a}_{i})-f(\mathbf{x}^{n}_{i})\|+\alpha,0)
    \label{eq:metricLoss}
\end{equation}

Starting from such a baseline predictor (\eg, triplet), our key insight is that we can use the model's learned feature embedding to generate visual explanations, in the form of attention maps, for why the current input triplet satisfies the triplet criterion. We refer to this set of attention maps as \textsl{similarity attention}. Note that our idea of generating attention maps conditioned on the feature embedding is different from existing work that relies on categorical classification modules \cite{GradCAM_ICCV17}, attribute classification modules \cite{whymatch_2019}, or nearest-neighbor-search databases of training gradients  \cite{Chen_WACV20}. In our case, we are not limited by these requirements, instead computing a score directly from the feature vectors to generate the explanations. A crucial advantage with this proposed strategy is the resulting flexibility and generality in visually explaining any feature embedding network, as discussed in the next section.

Given a triplet $(\mathbf{x}^{a},\mathbf{x}^{p},\mathbf{x}^{n})$, we first extract feature vectors $f(\mathbf{x}^{a})$, $f(\mathbf{x}^{p})$, and $f(\mathbf{x}^{n})$ (denoted $\mathbf{f}^{a}$, $\mathbf{f}^{p}$, and $\mathbf{f}^{n}$ respectively going forward, all normalized to have unit $l_{2}$ norm). A perfectly trained triplet similarity model must result in $\mathbf{f}^{a}$, $\mathbf{f}^{p}$, and $\mathbf{f}^{n}$ satisfying the triplet criterion. Under this scenario, local differences in the image space will roughly correspond to proportional differences in the feature space, and hence there must exist some feature dimensions contributing the most to the triplet criterion being respected. Our idea is to generate attention maps conditioned on these feature dimensions. To this end, we compute the absolute differences and construct the weight vectors $\mathbf{w}^{p}$ and $\mathbf{w}^{n}$ as $\mathbf{w}^{p}=\mathbf{1}-|\mathbf{f}^{a}-\mathbf{f}^{p}|$ and $\mathbf{w}^{n}=|\mathbf{f}^{a}-\mathbf{f}^{n}|$. With $\mathbf{w}^{p}$, we seek to highlight the feature dimensions that have a small absolute difference value (\eg, for those dimensions $t$, $\mathbf{w}^{p}_{t}$ will be closer to 1), whereas with $\mathbf{w}^{n}$ we seek to highlight the feature dimensions with large absolute differences. Given $\mathbf{w}^{p}$ and $\mathbf{w}^{n}$, we construct a single weight vector $\mathbf{w}=\mathbf{w}^{p} \odot \mathbf{w}^{n}$, where $\odot$ denotes the element-wise product operation. With $\mathbf{w}$, we obtain a higher weight on feature dimensions that have a high value in both $\mathbf{w}^{p}$ and $\mathbf{w}^{n}$. 

To further understand this intuition, let us consider a simple example. If the first feature dimension $f^{a}(1)=0.80$ and $f^{p}(1)=0.78$, then this first dimension is important for the anchor to be close to the positive. In this case, the first dimension of the corresponding weight vector $w^{p}(1)=(1-|0.80-0.78|)=0.98$, which is a high value, quantifying the importance of this particular feature dimension for the anchor and positive to be close. Similarly, if $f^{a}(5)=0.99$ and $f^{n}(5)=0.01$, this $5^{th}$ feature dimension of the corresponding weight vector would be $w^{n}(5)=|0.99-0.01|=0.98$, which is a high value, quantifying the importance of this particular feature dimension for the anchor and negative to be far apart. We identify all such high-value, or important, dimensions common across both $\textbf{w}^{p}$ and $\textbf{w}^{n}$ with the single weight vector $\mathbf{w}$. In other words, we focus on elements that contribute the most to the positive feature pair being close and the negative feature pair being further away (\ie, contributing to the triplet criterion) and then determine attention maps. 

To obtain the attention maps, we first calculate the dot product of $\mathbf{w}$ with each of $\mathbf{f}^{a}$, $\mathbf{f}^{p}$, and $\mathbf{f}^{n}$ to get the sample scores $s^{a}=\mathbf{w}^{T}\mathbf{f}^{a}$, $s^{p}=\mathbf{w}^{T}\mathbf{f}^{p}$, and $s^{n}=\mathbf{w}^{T}\mathbf{f}^{n}$ for each image in the triplet $(\mathbf{x}^{a},\mathbf{x}^{p},\mathbf{x}^{n})$. We then calculate the gradients of these scores with respect to the image's convolutional feature maps to get its attention map. Specifically, given a score $s^{i}, i\in \{a,p,n\}$, the attention map $\mathbf{M}^{i} \in \mathbb{R}^{m\times n}$ is determined as:
\begin{equation}
  \mathbf{M}^{i}=\text{ReLU}\left(\sum_k \alpha_{k}\mathbf{A}_{k}\right)  
  \label{eq:simAtt}
\end{equation}
where $\mathbf{A}_{k} \in \mathbb{R}^{m\times n}$ is the $k^{th} (k=1,\ldots,c)$ convolutional  feature channel (from an intermediate layer) of the feature map $\mathbf{A} \in \mathbb{R}^{m\times n \times c}$ and $\alpha_{k}=\text{GAP}\left(\frac{\partial s^{i}}{\partial \mathbf{A}_{k}}\right)$, while  \text{GAP} refers to global average pooling.

\subsubsection{Extensions to other architectures}
\label{sec:extensionSimAttention}
Our method is not limited to triplet CNNs and is extensible to other kinds of similarity architectures. For a Siamese model, the inputs are pairs $(\mathbf{x}^{1},\mathbf{x}^{2})$. Given their feature vectors $\mathbf{f}^{1}$ and $\mathbf{f}^{2}$, we compute the weight vector $\mathbf{w}$ in the same way as the triplet scenario. If $\mathbf{x}^{1}$ and $\mathbf{x}^{2}$ belong to the same class, $\mathbf{w}=\mathbf{1}-|\mathbf{f}^{1}-\mathbf{f}^{2}|$, otherwise, $\mathbf{w}=|\mathbf{f}^{1}-\mathbf{f}^{2}|$. With $\mathbf{w}$, and the sample scores $s^{1}=\mathbf{w}^{T}\mathbf{f}^{1}$ and $s^{2}=\mathbf{w}^{T}\mathbf{f}^{2}$, we compute attention maps $\mathbf{M}^{1}$ and $\mathbf{M}^{2}$ for $\mathbf{x}^{1}$ and $\mathbf{x}^{2}$ respectively using Equation~\ref{eq:simAtt}.

For a quadruplet model, the inputs are quadruplets $(\mathbf{x}^{a},\mathbf{x}^{p},\mathbf{x}^{n1},\mathbf{x}^{n2})$, where $\mathbf{x}^{p}$ is the positive sample and $\mathbf{x}^{n1}$ and $\mathbf{x}^{n2}$ are negative samples with respect to the anchor $\mathbf{x}^{a}$. Here, we compute the three difference feature vectors $\mathbf{f}^{1}=|\mathbf{f}^{a}-\mathbf{f}^{p}|$, $\mathbf{f}^{2}=|\mathbf{f}^{a}-\mathbf{f}^{n1}|$, and $\mathbf{f}^{3}=|\mathbf{f}^{a}-\mathbf{f}^{n2}|$. Following the intuition described in the triplet case, we compute the difference weight vectors as $\mathbf{w}^{1}=1-\mathbf{f}^{1}$ for the positive pair and $\mathbf{w}^{2}=\mathbf{f}^{2}$ and $\mathbf{w}^{3}=\mathbf{f}^{3}$ for the two negative pairs. The overall weight vector $\mathbf{w}$ is then computed as the element-wise product of the three individual weight vectors: $\mathbf{w}=\mathbf{w}^{1} \odot \mathbf{w}^{2} \odot \mathbf{w}^{3}$. With $\mathbf{w}$, and the sample scores $s^{a}=\mathbf{w}^{T}\mathbf{f}^{a}$, $s^{p}=\mathbf{w}^{T}\mathbf{f}^{p}$, $s^{n1}=\mathbf{w}^{T}\mathbf{f}^{n1}$, and $s^{n2}=\mathbf{w}^{T}\mathbf{f}^{n2}$, we use Equation~\ref{eq:simAtt} to obtain the four attention maps $\mathbf{M}^{a}$, $\mathbf{M}^{p}$, $\mathbf{M}^{n1}$, and $\mathbf{M}^{n2}$.

For other models trained with custom metric learning losses, all one has to do is compute feature vectors. The weight vector and sample scores, and hence attention maps, can then be easily determined as above. Given this dependence on just the feature embedding (obtainable universally for any CNN-based similarity architecture), our method is applicable to generic similarity models.

\subsection{Learning with similarity mining}
\label{sec:SA_mining}
With our proposed method, we can generate attention maps to explain the reasoning for a similarity model's predictions. However, we note all operations leading up to the attention map $\mathbf{M}^{i}$ in Section~\ref{sec:SA} are differentiable and thus we can use the generated attention maps to further bootstrap the training process. To this end, we propose a new learning objective called \textsl{similarity mining}. The goal of similarity mining is to facilitate the complete discovery of local image regions that the model deems necessary to satisfy the similarity criterion. 

Given the three attention maps $\mathbf{M}^{i}, i\in \{a,p,n\}$ (in the triplet case), we upsample them to be the same size as the input image and perform soft-masking, producing masked images that exclude pixels corresponding to high-response regions in the attention maps. This is realized as: $\hat{\mathbf{x}}=\mathbf{x} \odot (\mathbf{1}-\Sigma({\mathbf{M}}))$, where $\Sigma(\mathbf{Z})=\text{sigmoid}(\alpha(\mathbf{Z}-\beta))$ (all element-wise operations and $\alpha$ and $\beta$ are constants pre-set by cross validation). These masked images are then fed back to the same encoder of the triplet model to obtain the feature vectors $\mathbf{f}^{a*}$, $\mathbf{f}^{p*}$, and $\mathbf{f}^{n*}$.  Our proposed similarity mining loss $L_{sm}$ can then be expressed as: 
\begin{equation}
L_{sm}=\Bigl\lvert\|\mathbf{f}^{a*}-\mathbf{f}^{p*}\|-\|\mathbf{f}^{a*}-\mathbf{f}^{n*}\|\Bigr\rvert
\label{eq:simMining}
\end{equation}
where $\|\mathbf{t}\|$ represents the Euclidean norm of the vector $\mathbf{t}$. The intuition here is that by minimizing $L_{sm}$, the model has difficulties in predicting whether the input triplet would satisfy the triplet condition. This is because as $L_{sm}$ gets smaller, the model will have exhaustively discovered all possible local regions in the triplet, and erasing these regions (via soft-masking above) will leave no relevant features available for the model to predict that the triplet satisfies the criterion.

\begin{figure*}[!h]
	\centering
	\includegraphics[draft=false,width=1\linewidth]{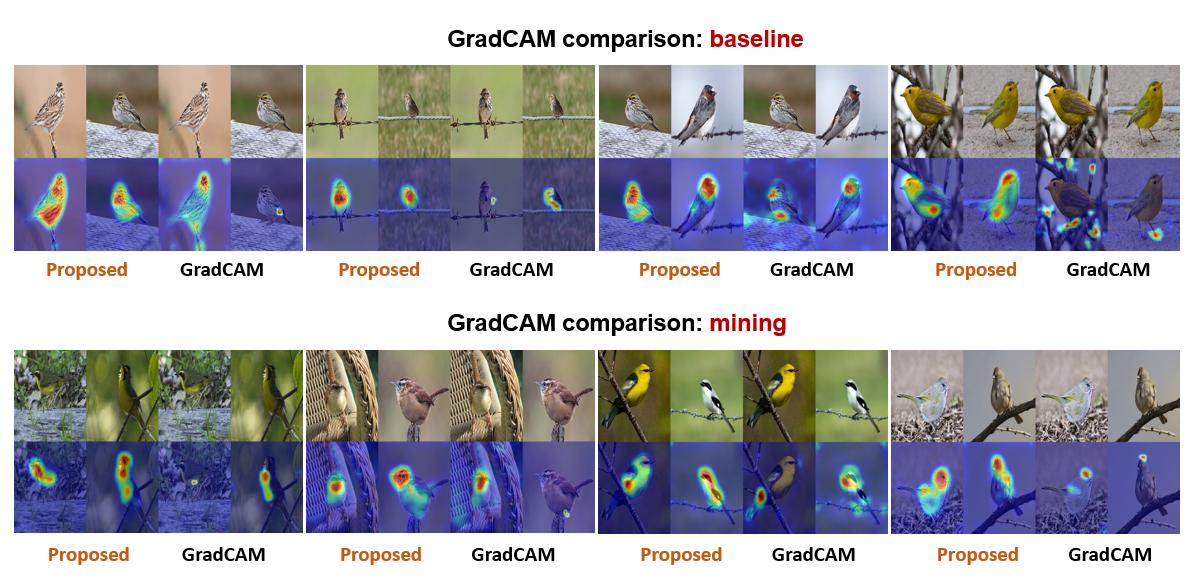}%
	\caption{Proposed similarity attention vs. GradCAM: we show four examples each with baseline and mining.} 
	\label{fig:VsGradCAM}
\end{figure*}

\subsubsection{Extensions to other architectures}
Like similarity attention, similarity mining is also extensible to other similarity learning architectures. For a Siamese model, given the two attention maps $\mathbf{M}^{1}$ and $\mathbf{M}^{2}$, we perform the soft-masking operation described above to obtain the masked images and their feature vectors $\mathbf{f}^{1*}$ and $\mathbf{f}^{2*}$. The similarity mining objective in this case attempts to maximize the distance between $\mathbf{f}^{1*}$ and $\mathbf{f}^{2*}$, \ie, 

\begin{equation}
L_{sm}=-|\mathbf{f}^{1*}-\mathbf{f}^{2*}| 
\end{equation}
The intuition here is that we seek to get the model to a state where after erasing, the model can no longer predict that the data pair belongs to the same class. This is because as $L_{sm}$ gets smaller, the model will have exhaustively discovered all corresponding regions that are responsible for the data pair to be predicted as belonging to the same class (\ie, low feature space distance), and erasing these regions (via soft-masking) will result in a larger feature space distance between the positive samples. 

For a quadruplet model, using the four attention maps, we compute the feature vectors $\mathbf{f}^{a*}$, $\mathbf{f}^{p*}$, $\mathbf{f}^{n1*}$, and $\mathbf{f}^{n2*}$ using the same masking strategy above. We then consider the two triplets $T_{1}=(\mathbf{f}^{a*},\mathbf{f}^{p*},\mathbf{f}^{n1*})$ and $T_{2}=(\mathbf{f}^{a*},\mathbf{f}^{p*},\mathbf{f}^{n2*})$ in constructing the similarity mining objective as

\begin{equation}
L_{sm}=L_{sm}^{T_{1}}+L_{sm}^{T_{2}}   
\end{equation}
where $L_{sm}^{T_{1}}$ and $L_{sm}^{T_{2}}$ correspond to Equation~\ref{eq:simMining} evaluated for $T_{1}$ and $T_{2}$ respectively.

\subsection{Overall training objective}
We train similarity models with both the traditional similarity/metric learning objective $L_{ml}$ (\eg, contrastive, triplet, etc.) as well as our proposed similarity mining objective $L_{sm}$. With $L_{ml}$ (triplet variant) defined in Equation~\ref{eq:metricLoss} and $L_{sm}$ (triplet variant) defined in Equation~\ref{eq:simMining}, our overall training objective is:
\begin{equation}
    L=L_{ml}+\gamma L_{sm}
\end{equation}
where $\gamma$ is a weight factor controlling the relative importance of $L_{ml}$ and $L_{sm}$. See Fig.~\ref{fig:pipeline} for a summary.

\begin{figure*}[!h]
	\centering
	\includegraphics[draft=false,width=.95\linewidth]{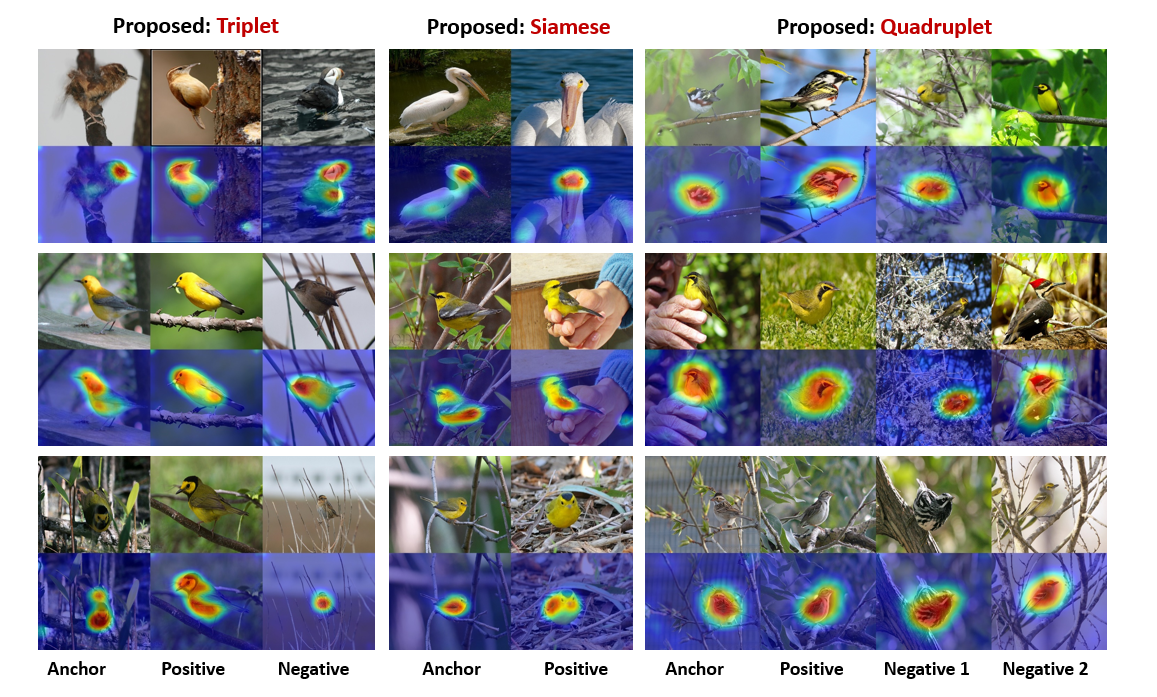}%
	\caption{Proposed similarity attention for triplet, Siamese, and quadruplet models.} 
	\label{fig:Generality_demo}
\end{figure*}

\begin{figure*}[!h]
	\centering
	\subfloat[Triplet attention maps on CUB dataset for model trained on CUB (left) and CARS (right).]{\includegraphics[draft=false,width=1\linewidth]{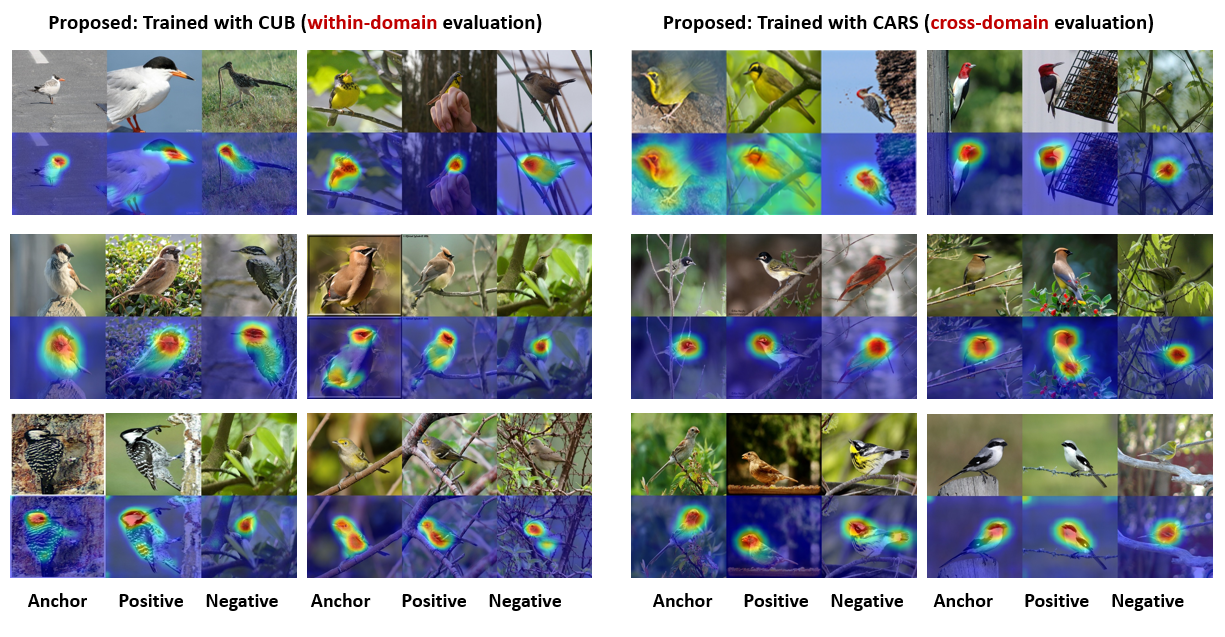}}\\%
	\subfloat[Triplet attention maps on CARS dataset for model trained on CARS (left) and CUB (right).]{\includegraphics[draft=false,width=1\linewidth]{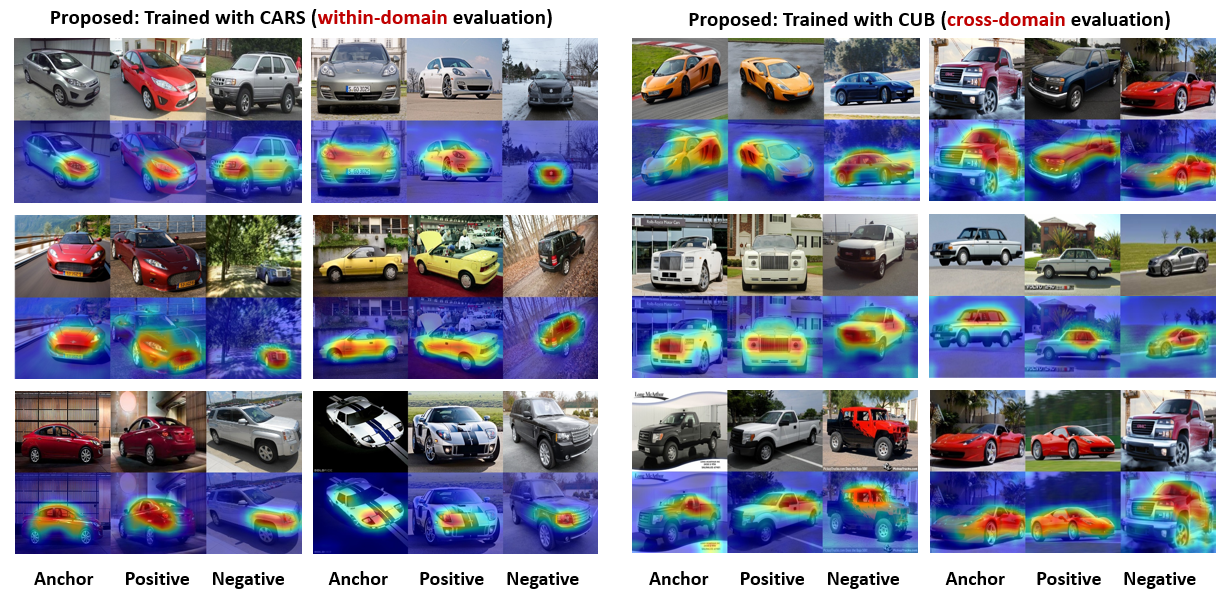}}
	\caption{Triplet attention maps on (a) CUB and (b) CARS with our proposed method for models trained with CUB and CARS.} 
	\label{fig:cubcarresults}
\end{figure*}

\section{Experiments and Results}
We conduct experiments on three different tasks: image retrieval (Sec.~\ref{sec:retr}), person re-identification (Sec.~\ref{sec:reid}), and one-shot semantic segmentation (Sec.~\ref{sec:segm}). We use a pretrained ResNet50 as our base architecture and implement all our code in Pytorch. All implementation details are available in the supplemental material. 

\subsection{Image retrieval}
\label{sec:retr}
We conduct experiments on the CUB200 (``CUB'') \cite{wah2011caltech}, Cars-196 (``CARS'') \cite{krause20133d} and Stanford Online Products (``SOP'') \cite{LiftStruct_CVPR15} datasets. We first discuss qualitative results, \ie, similarity attention maps, obtained with our method. In Figure~\ref{fig:VsGradCAM}, we compare our similarity attention with those of GradCAM for a Siamese model (please note four pairs in each of the ``baseline'' and ``mining'' sections). As noted in Section~\ref{sec:intro}, the GradCAM attention maps were obtained by training the model with a classification loss (we use binary cross entropy, $L_{bce}$) in addition to the standard metric learning loss (so $L_{ml}+\gamma L_{bce}$ for GradCAM whereas our proposed attention uses only $L_{ml}$). One can note from Fig.~\ref{fig:VsGradCAM} (``baseline'') that our similarity attention maps more accurately capture the corresponding regions in these images, with high response regions helping explain why the pairs of images are similar. This is not the case with GradCAM, with mostly non-corresponding regions being highlighted. 
The same observations can be made with models trained with our mining loss as well (``mining'' in Fig.~\ref{fig:VsGradCAM}), but now using $L_{bce}$ to compute attention before calculating the loss in Equation~\ref{eq:simMining} ($L_{bce-sm}$ to highlight GradCAM attention mining instead of our proposed similarity mining). In other words, the model (used to compute GradCAM attention maps) is trained with $L_{ml}+\gamma_{1} L_{bce}+\gamma_{2} L_{bce-sm}$. These results suggest that simply adding a classification loss term to compute attention maps, as in GradCAM, is not enough. 

We further show a quantitative comparison (testing results on CUB) in Table \ref{table:comp_gradcam}. In computing all numbers, we follow the protocol of Wang \etal \cite{msLoss_CVPR19} and use the standard Recall@K (R-K) metric \cite{msLoss_CVPR19}. Note that ``Baseline'', ``Baseline + BCE'' and ``Baseline + BCE + mining'' here indicate training with only the metric learning loss $L_{ml}$, with metric learning and BCE loss $L_{ml} + \gamma L_{bce}$, and with $L_{ml}+\gamma_{1} L_{bce}+\gamma_{2} L_{bce-sm}$ respectively. One can note from Table \ref{table:comp_gradcam} that adding a classification (BCE) loss (just so we can compute GradCAM attention) reduces the performance of the baseline Siamese model trained with only the metric learning loss (rank-1 performance drops by 1.7\%). With the attention maps obtained in this manner not as accurate as desired (see Fig. \ref{fig:VsGradCAM}), the results of the model with the mining objective deteriorate further (notice a 0.8\% R-1 drop for ``Baseline+BCE+mining'' compared to ``Baseline+BCE''). On the other hand, with our proposed method, called SAM (similarity attention and mining), we not only obtain quantitative improvements over the baseline Siamese model, we are also able to produce more meaningful similarity attention maps.

\begin{table}[h!]
\centering
\scalebox{1}{
\begin{tabular}{c|c|c|c}
\hline
Method &R-1 & R-2 &R-4 \\
    \hline\hline
    Baseline ~($L_{ml}$) & 65.9 & 77.5 & 85.8 \\
    \hline
    Baseline + BCE~
    ($L_{ml} + \gamma L_{bce}$) & 64.2 & 75.8 & 85.1 \\
    \hline
    \Centerstack{Baseline + BCE + mining \\ ($L_{ml}+\gamma_{1} L_{bce}+\gamma_{2} L_{bce-sm}$)} & 63.4 & 75.6 & 84.5 \\
    \hline\hline
    \textbf{Proposed SAM} ($L_{ml}+\gamma L_{sm}$) & \textbf{68.3} & \textbf{78.3} & \textbf{86.3} \\
    \hline
\end{tabular}}
\caption{Comparison with GradCAM on CUB. All numbers in $\%$. Here $\gamma$, $\gamma_{1}$, and $\gamma_{2}$ are set to 0.25.}
\label{table:comp_gradcam}
\end{table}

\begin{figure}[!h]
\centering
\includegraphics[draft=false,width=\linewidth]{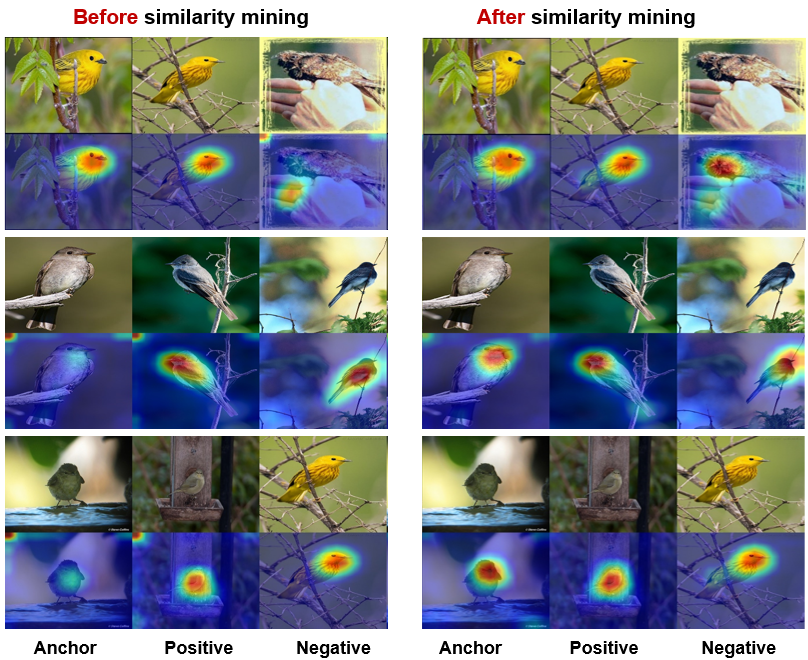}
\caption{Impact of similarity mining on triplet attention.}
\label{fig:mining}
\end{figure}

\begin{table*}[h!]
\centering
\scalebox{1}{
\begin{tabular}{c|c|c|c|c}
\hline
Arch. &Type &R-1 & R-2 &R-4 \\
    \hline\hline
    \multirow{2}{*}{Siamese}&Baseline & 65.9&77.5&\textbf{85.8}\\
    &\textbf{SAM} &\textbf{67.7}&\textbf{77.8}&85.5\\
    \hline
    \multirow{2}{*}{Triplet}&Baseline &66.4&78.1&85.6\\
    &\textbf{SAM} &\textbf{68.3}&\textbf{78.9}&\textbf{86.5}\\
    \hline
    \multirow{2}{*}{Quadruplet}&Baseline & 64.7&75.6&\textbf{85.2}\\
    &\textbf{SAM} & \textbf{66.4}&\textbf{77.0}&\textbf{85.2}\\
    \hline
\end{tabular} 
} \hspace{1em}
		\setlength{\extrarowheight}{.2em}
		\begin{tabular}{p{2.4cm}|cc|cc|cc}
		    \hline
		     & \multicolumn{2}{c|}{CUB} & \multicolumn{2}{c|}{Cars}  & \multicolumn{2}{c}{SOP}\\
			\cline{2-7}
			& R-1 & R-2 & R-1 & R-2 & R-1 & R-1k\\
			\hline\hline
			Lifted \cite{LiftStruct_CVPR15} & 47.2 & 58.9 & 49.0 & 60.3 & 62.1 & 97.4\\
			N-pair \cite{Npairmc_NIPS16} & 51.0 & 63.3 & 71.1 & 79.7 & 67.7 & 97.8\\
			P-NCA \cite{ProxyNCA_ICCV17} & 49.2 & 61.9 & 73.2 & 82.4 & 73.7 & -\\
			HDC \cite{HDC_ICCV2017} & 53.6 & 65.7 & 73.7 & 83.2 & 69.5 & 97.7\\
			BIER \cite{BIER_ICCV17} & 55.3 & 67.2 & 78.0 & 85.8 & 72.7 & 98.0\\
			ABE \cite{AttentionBasedEF_ECCV18} & 58.6 & 69.9 & 82.7 & 88.8 & 74.7 & 98.0 \\
			MS \cite{msLoss_CVPR19} & 65.7 & 77.0 & 84.1 & 90.4 & \textbf{78.2} & 98.7\\ 
			HDML \cite{zheng2019hardness} & 53.7 & 65.7  &79.1 & 89.7 & 68.7 & -\\
			DeML \cite{chen2019hybrid} & 65.4 & 75.3 & 86.3 & 91.2 & 76.1 & 98.1 \\
			GroupLoss \cite{Elezi_2020_ECCV} & 66.9 & 77.1 & \textbf{88.0} & \textbf{92.5} & 76.3 & - \\
			MS+SFT \cite{SFT_eccv2020} & 66.8 & 77.5 & 84.5 & 90.6 & 73.4 & - \\
			$\text{DRO-KL}_{M}$ \cite{qi2019simple_eccv2020} & 67.7 & 78.0 & 86.4 & 91.9 & - & - \\
			\hline
 			\textbf{SAM}  & \textbf{68.3}&\textbf{78.9} & 86.3 & 91.4 & 77.9 & \textbf{98.8}\\
			\hline
	\end{tabular}
\caption{\textbf{Left}: Ablation on CUB. \textbf{Right}: Results on CUB, CARS, and SOP. All numbers in $\%$.}
\label{table:ablationSOTAML}
\end{table*}

Next, to demonstrate generality, in Fig.~\ref{fig:Generality_demo}, we present our similarity attention maps for various architectures: Siamese, triplet, and quadruplet. In each of these cases, SAM is able to highlight intuitively satisfying regions (\eg, face/neck regions in the triplet case), providing visual evidence for similarity (in the Siamese case) or why they satisfy the training criterion (triplet or quadruplet loss). 

To further demonstrate the generalizability in obtaining these visual explanations, in Fig.~\ref{fig:cubcarresults}, we show within- and cross-domain triplet attention maps (for (a) CUB and (b) CARS testing data). For example, in Fig.~\ref{fig:cubcarresults}(a), we show attention maps with CUB testing data with model trained on CUB (left) and CARS (right). SAM is generally able to highlight intuitively satisfying corresponding regions with both  models. For instance, in the top-right triplet (for birds) of Fig.~\ref{fig:cubcarresults}(a) (cross-domain evaluation), the region around the face is what makes the second bird image similar, and the third bird image dissimilar, to the first (anchor) bird image. These results provide evidence for the generalizability of our generated visual explanations, with even a model not trained on relevant data (trained on CARS but tested on CUB or trained on CUB but tested on CARS) able to discover local regions contributing to the final decision. Here, we also show the impact of our proposed similarity mining loss on the generated attention maps in Fig.~\ref{fig:mining} (left triplet: baseline $L_{ml}$, right triplet: proposed $L_{ml}+\gamma L_{sm}$). We clearly see that the proposed $L_{sm}$ results in more exhaustive and accurate discovery of local regions, further demonstrating its impact in improving model performance. Additional qualitative results are available in the supplemental material. 

Finally, we report quantitative performance compared to competing metric learning methods (we follow the same protocols and metrics as noted above). We begin with ablation results to demonstrate performance gains achieved by our proposed similarity mining loss of Section~\ref{sec:SA_mining}. In Table~\ref{table:ablationSOTAML} (Left), we show both baseline (trained only with $L_{ml}$) and our results with the Siamese, triplet, and quadruplet architectures (trained with $L_{ml}+\gamma L_{sm}$). One can note that SAM consistently improves the baseline performance across all three architectures. Next, we compare this performance with contemporary methods in Table~\ref{table:ablationSOTAML} (Right), where we note SAM (with the triplet variant) is quite competitive, with R-1 performance improvement of $0.6\%$ on CUB, matching (with DRO-KL$_{M}$) R-1 and slightly better R-1k performance (w.r.t. MS \cite{msLoss_CVPR19}) on SOP. We emphasize that in addition to these competitive numbers, SAM is also able to provide reasoning in the form of attention maps, as discussed above, unlike these competing methods.

\subsection{Person re-identification}
\label{sec:reid}
\begin{figure}[!h]
\centering
\includegraphics[draft=false,width=0.9\linewidth]{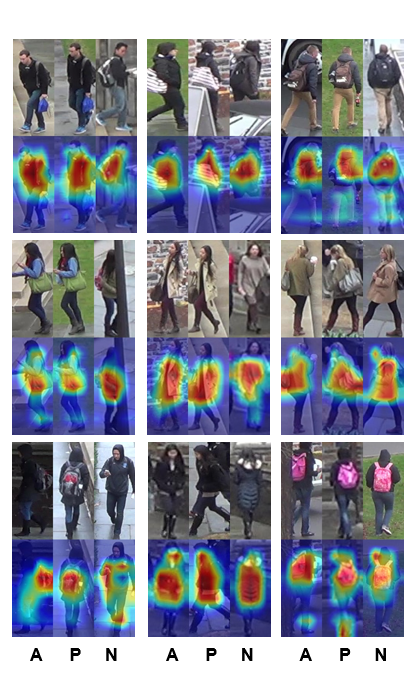}
\caption{Re-id attention maps.}
\label{fig:reidAttention}
\end{figure}

Since re-id \cite{zheng2016person,karanam2018systematic} is a special case of image retrieval, our method is certainly applicable, and we conduct experiments on CUHK03-NP detected (``CUHK-D'') and labeled (``CUHK-L'') \cite{li2014deepreid,zhong2017re} and Market-1501 (``Market") datasets, following the protocol in Sun \etal \cite{sunPCB_ECCV18}. We use the baseline architecture of Sun \etal \cite{sunPCB_ECCV18}, set $\gamma=0.2$ and train the model for 40 epochs with the Adam optimizer. We show attention maps with SAM in Fig.~\ref{fig:reidAttention} where we highlight (in red) image regions that the model reasons as being important for them to represent the same person (\eg, the white bag in the middle column). A key difference between our result and those in CASN \cite{zheng2019re} is that SAM does not need a BCE classification term to generate these attention maps. Furthermore, SAM does not make any re-id specific design choices (\eg, upright pose assumption for attention consistency in CASN \cite{zheng2019re}, hard attention in HA-CNN \cite{LiHACNN_CVPR18}, attentive feature refinement and alignment in DuATM \cite{si2018dual}) and is able to obtain meaningful visual explanations. We also report quantitative performance in Table~\ref{tab:reid}, where we note SAM achieves competitive performance: about $3\%$ rank-1 performance improvement on CUHK and close performance (94.4\% rank-1) to the best performing method (MGN) on Market.

\begin{table}[]
    \centering
    \scalebox{1}{
    \begin{tabular}{p{1.8cm}|cc|cc|cc}
		    \hline
		     & \multicolumn{2}{c|}{CUHK-D}& \multicolumn{2}{c|}{CUHK-L} & \multicolumn{2}{c}{Market} \\
			\cline{2-7}
			& R-1 & mAP & R-1 & mAP & R-1 & mAP \\
			\hline\hline
			SVD \cite{SVDnet_ICCV17} & 41.5 &37.3 & - & - & 82.3 & 62.1\\
			HA \cite{LiHACNN_CVPR18} & 41.7 & 38.6 & 44.4 & 41.0 & 91.2 & 75.7 \\
            DA \cite{si2018dual} & - & - & 41.5 & 37.3 & 91.4 & 76.6 \\
            DaRe \cite{DaRe_CVPR18} & 63.3 & 59.0 & 66.1 & 61.6 & 89.0 & 76.0 \\
			PCB \cite{sunPCB_ECCV18} & 63.7 & 57.5 & - & - & 93.8 & 81.6 \\
			CBN \cite{zhuang2020rethinking}  & - & - & - & - & 94.3 & 83.6 \\
 			MGN \cite{MGN_MM18}  & 66.8 & 66.0 & 68.0 & 67.4 &\textbf{95.7} & \textbf{86.9}\\
            CASN \cite{zheng2019re} & 71.5 & 64.4 & 73.7 & 68.0 & 94.4 & 82.8  \\
            GASM \cite{GASM_eccv2020}  & - & - & - & - & 95.3 & 84.7 \\
            \hline
            \textbf{SAM} & \textbf{74.5} & \textbf{67.5} & \textbf{76.3} & \textbf{70.5} & 94.4 & 83.1 \\
            \hline
	\end{tabular}
	\caption{Rank-1 and mAP evaluation of proposed approach on CUHK03-NP (detected and labeled) \cite{li2014deepreid,zhong2017re} and Market-1501 \cite{Market1501_ICCV15} datasets.}
	\label{tab:reid}}
\end{table}

\begin{figure*}[h!]
	\centering
	\includegraphics[draft=false,width=0.95\linewidth, height=9cm]{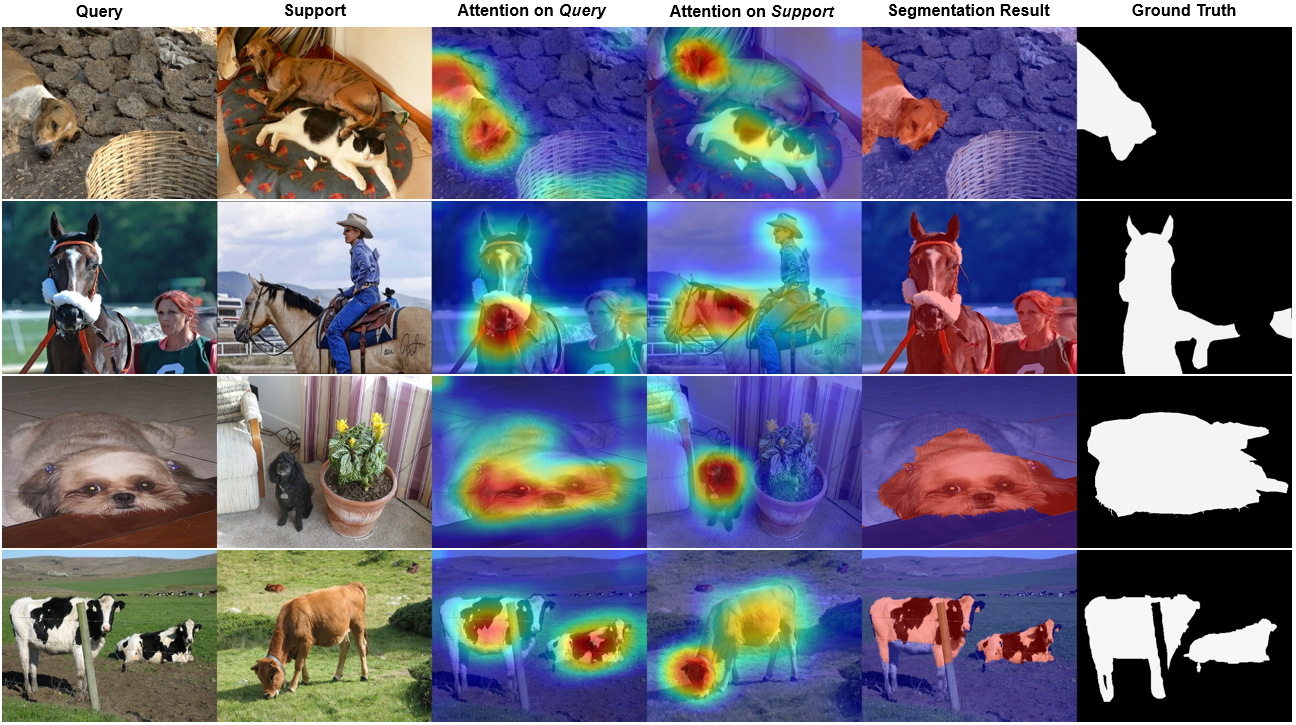}\\
	\includegraphics[draft=false,width=0.95\linewidth, height=4cm]{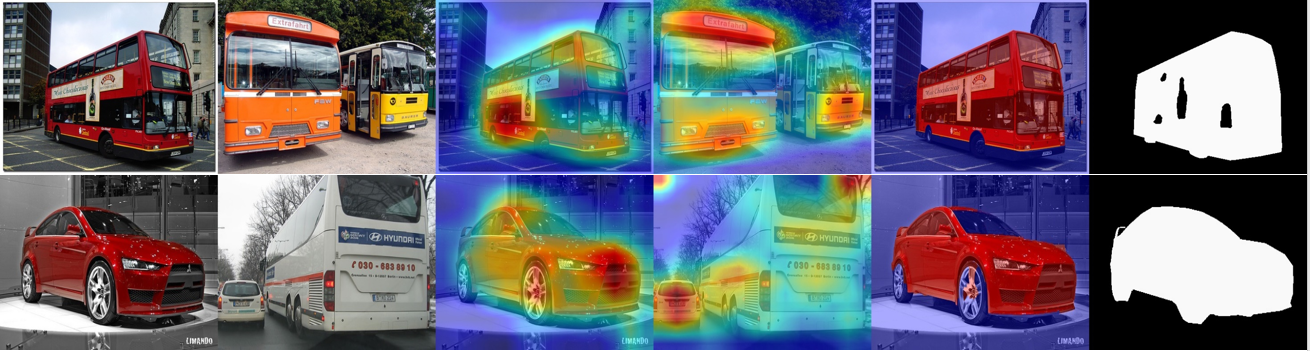}
	\caption{Qualitative one-way one-shot segmentation results from the PASCAL$-5i$ dataset.}
	\label{fig:onewayoneShotAttResults}
	\vspace{-0.5em}
\end{figure*}

\begin{figure*}[h!]
	\centering
	\subfloat[2-way 1-shot]{\includegraphics[draft=false,width=1.0\linewidth]{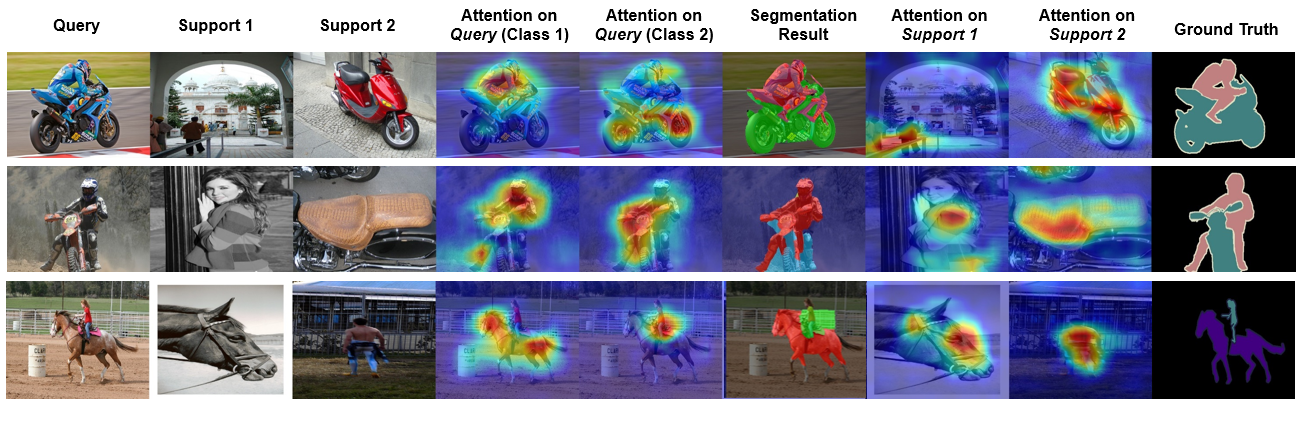}}\\
	\subfloat[2-way 5-shot]{\includegraphics[draft=false,width=1.0\linewidth]{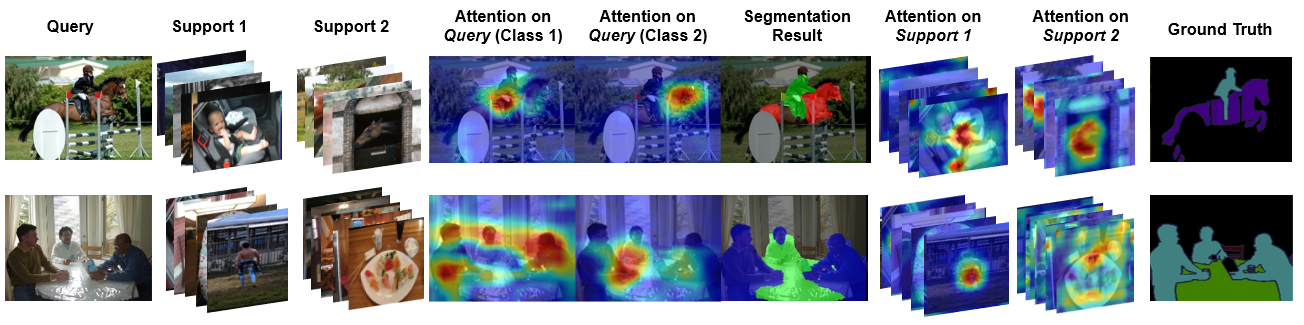}}\\
	\caption{Qualitative two-way one-shot and five-shot segmentation results from the PASCAL$-5i$ dataset.}
	\label{fig:twowayNShotAttResults}
	\vspace{-0.5em}
\end{figure*}

\begin{table*}[!h]
\centering
    \begin{minipage}[b]{0.47\hsize}
    \centering
    \begin{tabular}{c|c|c|c|c|c|c}
        \hline
        Methods & Label &$5^{0}$ &$5^{1}$ & $5^{2}$ & $5^{3}$ & Mean \\
        \cline{1-7}
        \hline\hline
        OSVOS \cite{caelles2017one}& Yes & 24.9 & 38.8 & 36.5 & 30.1 & 32.6 \\
        OSLSM \cite{shaban2017one} & Yes & 33.6 & \textbf{55.3} & 40.9 & 33.5 & 40.8 \\
        co-FCN \cite{rakelly2018conditional} & Yes & 36.7 & 50.6 & \textbf{44.9} & 32.4 & 41.1 \\
        \hline
        PAN-init \cite{PAN_ICCV19} & Yes & 30.8 & 40.7 & 38.3 & 31.4 & 35.3 \\
         \hline
        \textbf{SAM} & No & \textbf{37.9} & 50.3  & 44.4 & \textbf{33.8} & \textbf{41.6} \\ 
        \hline 
        \end{tabular}
    \end{minipage}
    \hfill
    \begin{minipage}[b]{0.45\hsize}
    \centering
    \scalebox{1}{
        \begin{tabular}{c|c|c|c}
           \hline
           Methods & Label & 1-shot & 5-shot \\
           \hline\hline
           PL \cite{Dong2018FewShotSS} & Yes & 39.7 & 40.3 \\ 
           PL+SEG \cite{Dong2018FewShotSS} & Yes & 41.9 & 42.6 \\ 
           PL+SEG+PT \cite{Dong2018FewShotSS} & Yes & 42.7 & 43.7 \\ 
           \hline
           \textbf{SAM} & No & \textbf{56.9} & \textbf{60.1} \\ 
           \hline
        \end{tabular}}
    \caption{1-way 1-shot (left) and 2-way 1-shot and 5-shot (right) binary-IOU results on PASCAL$-5^{i}$. All numbers in \%.}
    \label{tab:segResults}
    \end{minipage}%
\end{table*}

\subsection{One-shot semantic segmentation}
\label{sec:segm}
In the one-shot semantic segmentation task, we are given a test image and a pixel-level semantically labeled support image, and we are to semantically segment the test image. Given that we learn similarity predictors, we can use our model to establish correspondences between the test and the support images. One aspect that is particularly appealing with SAM is explainability, and the resulting similarity attention maps we generate can be used as cues to perform semantic segmentation. A key aspect to note here is that while existing work in one-shot semantic segmentation makes use of the available label map of the support image in inferring the label map of the test image, we do not use this support label in any capacity. Instead, we only rely on the test and support images, computing the two attention maps with SAM and then processing the test attention map to obtain the test label map. This way, we take a weakly-supervised approach to this problem, as opposed to the stronger supervision (by means of the extra support label map) of related methods \cite{caelles2017one,shaban2017one,rakelly2018conditional,Dong2018FewShotSS,PAN_ICCV19}. 

We use the PASCAL$-5^{i}$ dataset (``Pascal'') \cite{shaban2017one} for all experiments, following the same protocol as Shaban \etal \cite{shaban2017one}. Given a test image and the corresponding support image, we first use our trained Siamese similarity model to generate two similarity attention maps, one for each image. We then use the attention map for the test image as a cue to generate the final segmentation map using the GrabCut \cite{rother2004grabcut} algorithm. We call this the ``1-way 1-shot'' experiment. In the ``2-way 1-shot'' experiment, the test image has two objects of different classes and we are given two support images, one for each class. In this case, to generate results for object class 1, we use the support image for object class 1 as the positive image and the other support image as negative. Similarly, to generate results for object class 2, we use the support image for object class 2 as the positive image and other as negative. We then use our triplet similarity model to generate three attention maps (one for test, one for positive support and one for negative support) and process the test attention map as above (using it as cue with GrabCut) to generate the final segmentation map. The ``2-way 5-shot'' experiment is similar; the only difference is we now have five support images for each of the two classes (instead of one image as above). In this case, as above, we form five triplets (test, positive support, negative support) and generate five triplets of attention maps with our triplet similarity model. The resulting five test attention maps are then averaged to obtain the final test attention map, which is then used with GrabCut to generate the final segmentation map. 

We first show some qualitative results in Fig.~\ref{fig:onewayoneShotAttResults} and \ref{fig:twowayNShotAttResults} (left to right: test image, support image, test attention map, support image attention map, predicted segmentation mask, ground truth mask). In the first row of Fig.~\ref{fig:onewayoneShotAttResults} (1-way 1-shot), we see that, in the test attention map, our method is able to capture the ``dog'' region in the test image despite the presence of a ``cat'' in the support image, helping generate the final segmentation result. In Fig. \ref{fig:twowayNShotAttResults}, we see our method is able to disambiguate and segment out both the person and the bike/horse/table following the person and bike/horse/table categories present in the support images. Additional qualitative results (both 1-way and 2-way) are available in the supplemental material. 

Finally, we also report the 1-way (following the protocol of \cite{shaban2017one}) and 2-way (following the protocol of \cite{Dong2018FewShotSS}) meanIOU results in Table~\ref{tab:segResults}. Here we highlight several aspects. First, all these competing methods are specifically trained towards the one/few-shot segmentation task, whereas our model is trained for metric learning. Second, these techniques use the support image label map both during training and inference, whereas our method does not use this label data. Finally, these models are trained on Pascal, \ie, relevant data, whereas our model was trained on CUB and Cars, data irrelevant in this context. Despite these seemingly disadvantageous factors, SAM performs better than others in some cases and for the overall mean in the 1-way experiment, and substantially outperforms competing methods in the 5-way experiments. Finally, SAM also substantially outperforms the recently published PAN-init \cite{PAN_ICCV19} which also does not train on the Pascal data (so this is closer to our experimental setup), while however using the support label information during inference. These results demonstrate the potential of SAM in training similarity predictors that can generalize to data unseen during training and also to tasks for which the models were not originally trained.

\section{Summary and Future Work}

In this work, we considered the problem of visually explaining similarity models, i.e., producing a visual explanation (by means of attention maps) in addition to a scalar score representing the similarity of the input set of images. We discussed key issues with existing work, i.e., the need for classification-like architectures and class labels to generate attention maps, and proposed new techniques to address these issues. To remove the dependence on classification logits, we presented \textsl{similarity attention} to compute visual attention maps directly from the learned feature embedding. We showed how all the resulting operations are differentiable, leading to their use in enforcing trainable constraints, called \textsl{similarity mining}, for further improving model performance. This resulted in a new learning paradigm, called similarity attention and mining (SAM), that, during training, progressively improved the quality of the similarity attention maps by exhaustively discovering all local regions satisfying the metric learning objective. 

We demonstrated the flexibility and versatility of SAM in a variety of applications, \eg, generic image retrieval, person re-identification, and low-shot semantic segmentation. The segmentation experiments in particular showed that SAM can be used to address a variety of label propagation problems that need a first step of correspondence learning. Specifically, with our similarity attention, one can establish such correspondences in an unsupervised fashion, opening up new avenues for advances in zero- or few-shot learning. 

While our results are encouraging, more work needs to be done before we can put some of these ideas into real-world use. Below, we posit some directions for future research:

\begin{itemize}
    \item To realize the full potential of the ideas discussed here, we would need to apply them to a variety of unsupervised problem domains. As noted above, the attention maps resulting from image correspondences can be put to multipurpose use. For instance, given a reference image of an object of interest, we can detect instances of this object in a test image. Such low-shot object detection problems can find use in targeted retrieval for medical applications where a doctor can ``tag'' a certain region in the image under examination and retrieve relevant ``similar'' historical records \cite{akgul2011content,hegde2019similar}. While in many cases we would require application-specific design considerations (e.g., in medical applications, establishing correspondences so the attention map highlights the region of interest can be challenging since the underlying differences are usually subtle), our proposed similarity attention can act as a basis upon which extensions can be developed.
    \item One practical systems aspect our method has not addressed is the possibility that these similarity attention maps are counter-intuitive or wrong from a human's perspective. This is an important aspect that needs to be considered and addressed to be able to build human trust in the associated real-world system. One approach to resolve this issue would involve human-in-the-loop user studies and an active learning strategy for algorithm development. This would enable the creation of a cyclical framework where the algorithm can learn from, and improve itself with, human input while the human can (after establishing trust) rely on the algorithm for gaining new knowledge (\eg, discovering previously unrecognizable similarities in objects).  
    \item As a longer-term research challenge, our use of visual attention (i.e., heatmaps superimposed on images) as a means to explain similarity models can be construed as narrow if viewed from a broad sense of explainability and reasoning. Specifically, while our heatmaps can help explain the model's predictions, they are not enough to perform reasoning. As a concrete example, for an AI agent to be able to plan ahead to reach a certain goal, ``big-picture" reasoning and not just ``instantaneous" explainability (as in our proposed method) would be required. 
\end{itemize}

\bibliography{egbib}

\end{document}